\documentclass[10pt,letterpaper]{article}

\usepackage{cogsci}

\cogscifinalcopy 

\usepackage{amsmath}
\usepackage{algorithm2e}
\usepackage{graphicx}

\usepackage[small]{caption}
\usepackage{pslatex}
\usepackage{apacite}
\usepackage{float} 

\usepackage{color}
\newcommand\todo[1]{%
  \bgroup
  \hskip0pt\color{red}%
  TODO: #1%
  \egroup
}

\title{Compositional learning of functions in humans and machines}
 
\author{{\large \bf Yanli Zhou$^{1,3}$} \\ 
yanlizhou@nyu.edu \\\\
$^{1}$Center for Data Science\\
New York University
\And {\large \bf Brenden M. Lake$^{1,2}$} \\
brenden@nyu.edu\\\\
$^{2}$Department of Psychology\\
New York University
\And  {\large \bf Adina Williams$^{3}$} \\
adinawilliams@meta.com\\\\
$^{3}$Meta AI\\}

\begin{document}

\maketitle

\begin{abstract}

The ability to learn and compose functions is foundational to efficient learning and reasoning in humans, enabling flexible generalizations such as creating new dishes from known cooking processes. Beyond sequential chaining of functions, existing linguistics literature indicates that humans can grasp more complex compositions with interacting functions, where output production depends on context changes induced by different function orderings. Extending the investigation into the visual domain, we developed a function learning paradigm to explore the capacity of humans and neural network models in learning and reasoning with compositional functions under varied interaction conditions. Following brief training on individual functions, human participants were assessed on composing two learned functions, in ways covering four main interaction types, including instances in which the application of the first function creates or removes the context for applying the second function. Our findings indicate that humans can make zero-shot generalizations on novel visual function compositions across interaction conditions, demonstrating sensitivity to contextual changes. A comparison with a neural network model on the same task reveals that, through the meta-learning for compositionality (MLC) approach, a standard sequence-to-sequence Transformer can mimic human generalization patterns in composing functions.



\end{abstract}

\section{Introduction}

Humans are efficient and flexible learners, with the ability to infer an underlying function from exposure to just a few input-output examples. Humans are also adept in composing functions together in sequence: we can flexibly combine previously learned functions in new ways. For example, someone who knows how to chop vegetables and knows how to fry food can learn how to make fries, by chaining their chopping skills with their frying skills, even if they have never previously attempted that dish. Moreover, function composition skills emerge early in life, with children beginning to learn how to compose visual functions without explicit training at as early as 3.5 years of age \cite{Piantadosi2016b}. Mastery of these skills over time becomes a cornerstone for comprehending complex symbolic systems, fostering abstract thinking and aiding the acquisition of conceptual knowledge \shortcite{Curry1958, schonfinkel_1967, Piantadosi2012}. 

\begin{figure}[h]
\centering
\includegraphics[width=1.0\linewidth]{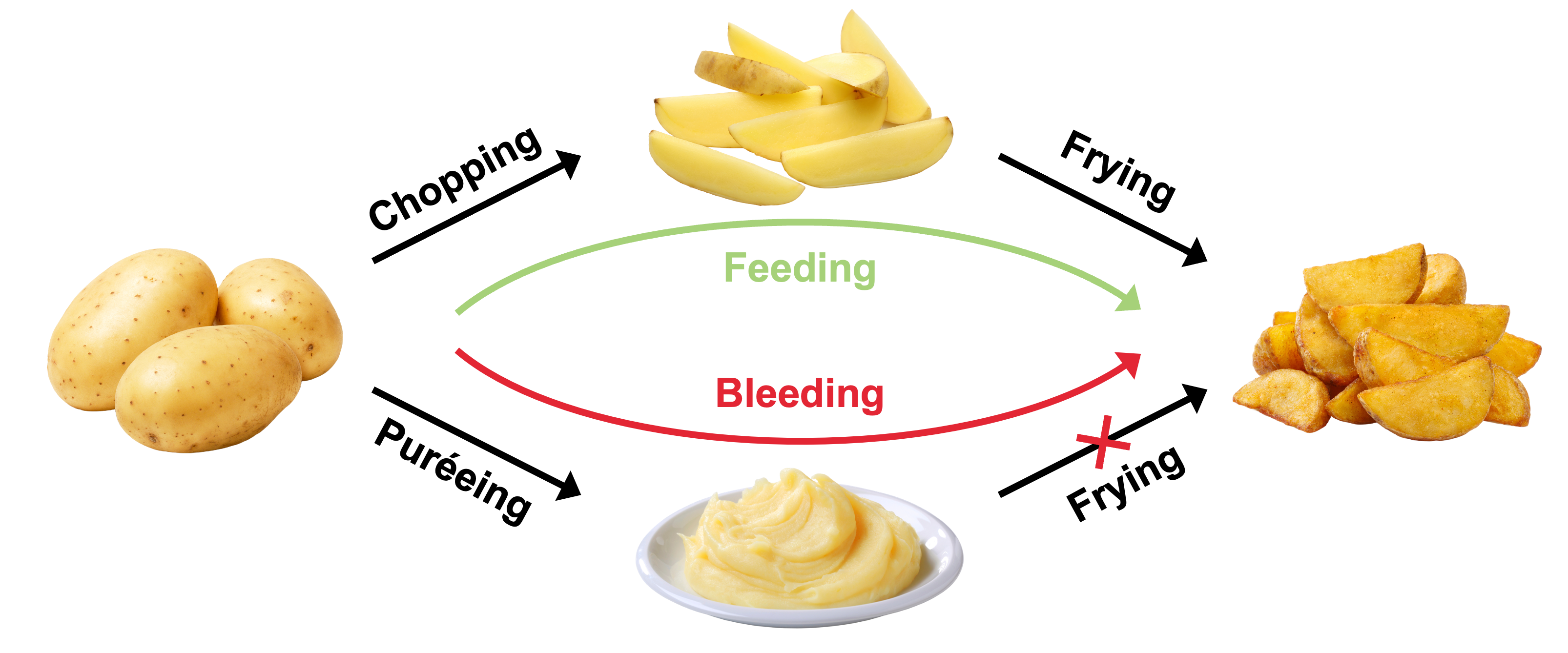}

\caption{Function interactions influence function compositions. The chopping operation provides context for frying while the pur\'{e}eing operation renders frying inapplicable.}
\label{fig:intro}
\vskip -1 em
\end{figure}

Beyond the sequential application of functions, humans can also track contextual changes that occur during function composition as a result of the order of operations. In the previous example, if the potatoes have been pur\'{e}ed, one would quickly realize that the frying operation can no longer take place, as the pur\'{e}eing function has transformed the potatoes into a state unsuitable for the second function to apply. Linguists have formalized different context-shifting phenomena into four types of function ordering \cite{Kiparsky1968}: \textit{feeding} describes scenarios in which the application of the first function creates the context for the second one to apply, as illustrated in the example when \textit{chopping} feeds the function \textit{frying} (Fig.\ref{fig:intro}, top); \textit{counter-feeding} describes the reverse order of application, with the context-creating function applied too late. For example, trying---and failing---to fry whole potatoes before chopping them. \textit{Bleeding} occurs when the context of the second function is removed by the first function's application, akin to the example where \textit{pur\'{e}eing} potatoes makes the unsuitable for later \textit{frying} (Fig.\ref{fig:intro}, bottom); counter-bleeding is its reverse, where the context-removal happens after the first function has successfully been applied (i.e., \textit{frying} does not prevent one from later \textit{pur\'{e}eing}). Together, we see that function composition emerges as a challenging learning program, requiring not only generalization to novel instances on the individual function level, but also to compositions of two functions with sensitivity to different interactions induced by various function orders.


Past research has investigated whether models can capture human-like compositional learning and generalization. For instance, \citeA{LakeBaroni2018} proposed the SCAN task to assess the compositional skills of neural network models. The authors, as well as in many subsequent investigations \shortcite{Bastings2018, Ruis2020, Valvoda2022}, found that neural networks continue to struggle with systematic generalizations despite recent AI advances. \shortciteA{liska2018} introduced a more explicit test of function composition, and found that only a small portion of tested recurrent neural networks converged to compositional solutions in computing the outputs of composite lookup tables. Previous studies have also evaluated model performance alongside empirical investigations, including explorations of how humans learn structured visual concepts by reasoning about how parts compose \shortcite{Overlan2017, zhou2024}. More recently, \citeA{Lake_Baroni_2023} showed that humans excel at composing functions in instruction learning, and demonstrated that a standard neural network-based model can be optimized to exhibit human-like behavior through a meta-learning procedure that encourages compositionality. Despite varying levels of modeling success reported in previous work on compositional function learning, none of them systematically studied the learning of functions and their interactions in humans and machines.

Towards this end, we proposed a learning paradigm suitable for testing both humans and models on their compositional function learning skills, with a special focus on the variety of function interactions. Extending the experimental framework from \citeA{Piantadosi2016b}, we investigated whether participants could learn visual functions as transformations of cartoon cars moving in and out of factory units with minimal input-output exposure. For each prompted input car, participants were evaluated on zero-shot function composition in feeding, counter-feeding, bleeding, and counter-bleeding conditions. Our experiment demonstrated that humans efficiently generalize from newly learned single functions to their compositions, achieving consistently high levels of accuracy across different orderings of visual functions, contrary to previous linguistic theories on human learning biases in function interactions. Our results suggested that humans are sensitive to contextual changes during function composition, generating different outputs based on the order of operations. 

Following \citeA{Lake_Baroni_2023}, we trained a neural network through meta-learning for compositionality (MLC) in order to learn functions and their compositional interactions. When comparing model performance directly to behavioral data on the same learning task, we found that a standard Transformer \shortcite{Vaswani2017}, without any explicit tooling for symbolic reasoning, can be trained to perform our function composition task at near-human accuracy. Additionally, we demonstrated by fine-tuning on behaviorally-informed data distributions \cite{mccoy2023, zhou2024}, model generations further captured nuanced error patterns observed in human-generated function outputs. Extending previous work on compositional function learning, our study presented empirical data on context and order-sensitive visual function composition. Side-by-side evaluations of human and machine performance illustrate that generic neural network models can \textit{learn-to-learn} function compositions through behaviorally-guided meta-learning.

\section{Behavioral experiment}

\begin{figure}[ht]
\centering
\includegraphics[width=1.0\linewidth]{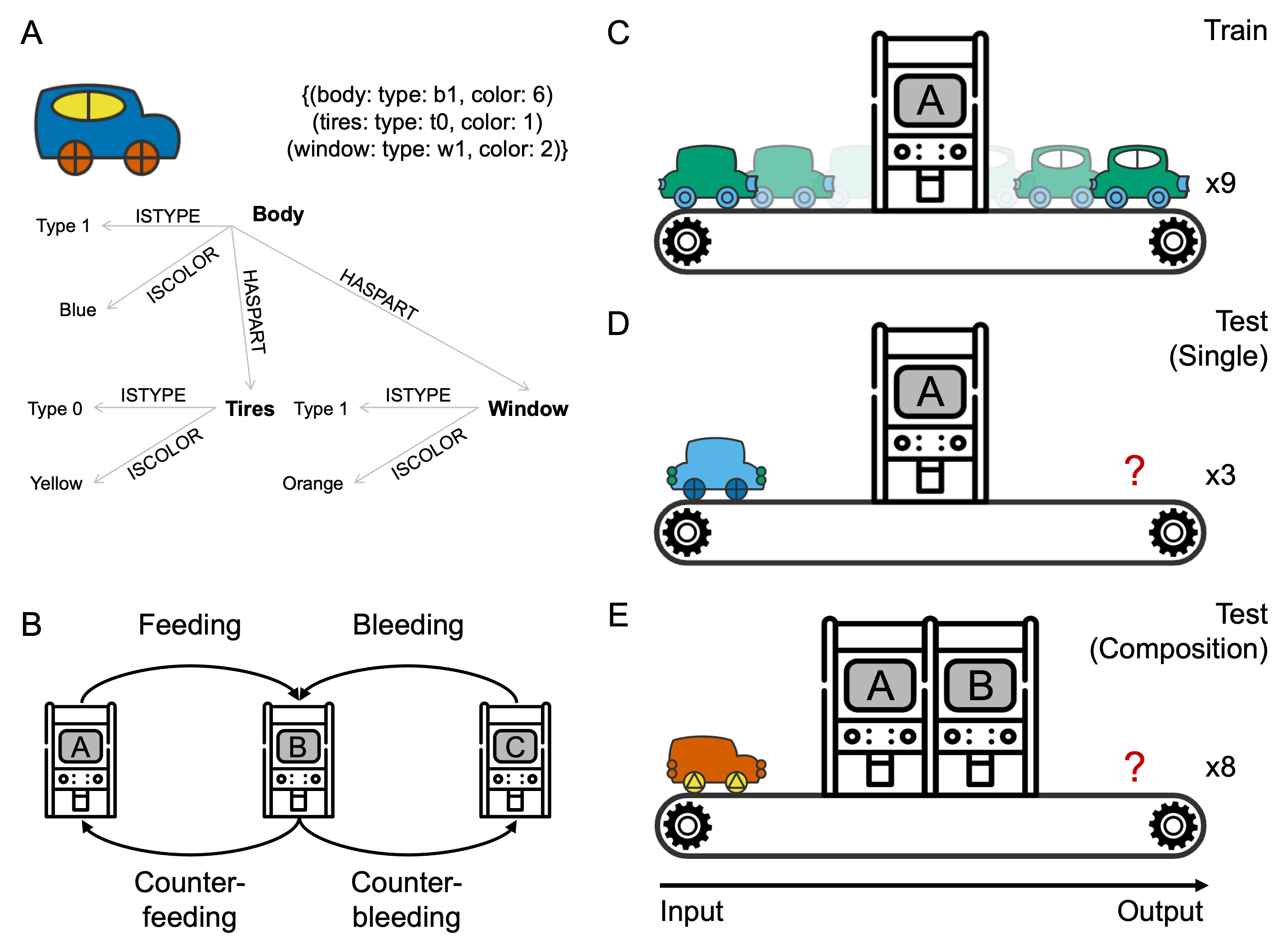}

\caption{Stimuli and experimental procedure. (A) Car stimuli are structured as trees: a car part like the window is a child node of the car body, and it has two nodes denoting its type and color. All functions are defined as tree edits that add, remove or modify nodes of the car tree. (B) Participants learn 3 functions $A$, $B$ and $C$ satisfying a set of 4 interaction relations. Arrows denote the order of function application (see concrete examples in Fig.\ref{fig:examples}). (C) During the training period, participants saw 9 cars moving through each factory unit representing a different function. If a car is a valid input to the function, it will come out of the unit with changes reflecting the underlying function. Cars remain unchanged if they are invalid inputs. (D) Participants were first asked to generate the correct output based on each prompted input car and factory unit. (E) For each of the interaction types, represented by the relevant factory units and the order of their application, participants were asked to generate the correct output car for 8 different input cars.}
\label{fig:task}
\vskip -1em
\end{figure}

\begin{figure}[h]
\centering
\includegraphics[width=1.0\linewidth]{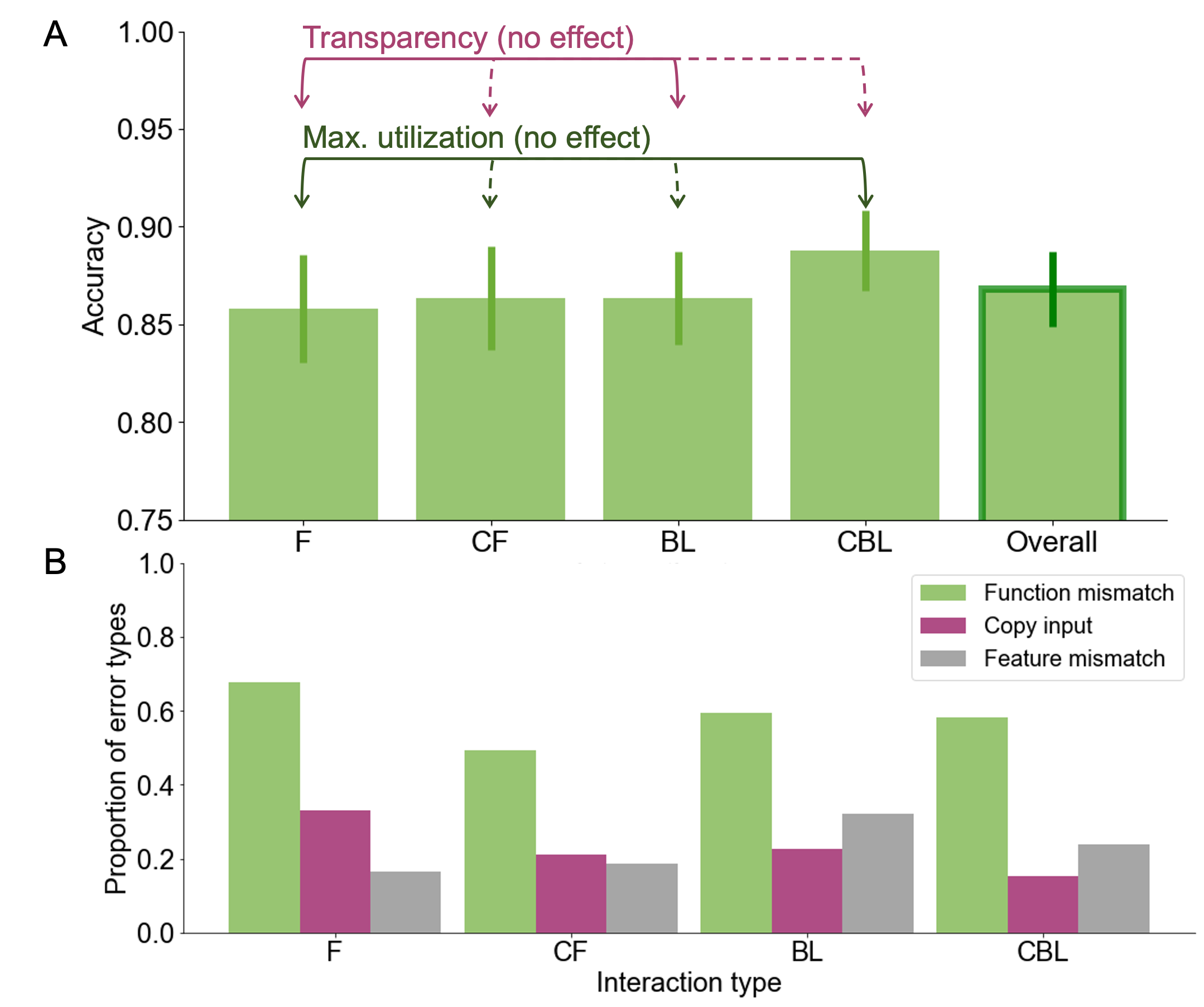}

\caption{Behavioral results. (A) Generation accuracy by interaction type. Performance did not differ significantly across different conditions. We also did not observe higher performance in F/CBL trials over CF/BL trials (maximum utilization bias), nor F/BL over CF/CBL trials (transparency bias). (B) Proportion of each error type over all incorrect generations by interaction type.}
\label{fig:human_res}
\vskip -1.5em
\end{figure}

 We retrofitted the task design introduced by \citeA{Piantadosi2016b} with an extended space of functions and features suitable for testing both humans and models in various interaction conditions. Specifically, participants took part in an online experiment titled ``car assembly game'' in which they played the role of production line workers in a car factory. The task was to assemble cars based on the knowledge of the different factory units installed on each assembly line. 

\subsubsection{Stimuli generation.} 

The main class of stimuli consisted of programmatically generated cartoon cars represented as tree-like objects (Fig.\ref{fig:task}A). Each car is a tree with its root node being the car body. The car body can have up to 3 different child nodes denoting the window, tires or lights respectively. Each car part also has two child nodes representing its type and color. There are 3 possible types for each car part, reflecting different part shapes, and 7 possible, colorblind safe colors, including a ``no color'' option. This defines a space of 375,000 distinct car objects. 

All functions are defined as tree operations on the car objects. For example, a function describing the action ``adding an oval shaped window to a car without any existing window" can be written as  

\begin{algorithm}
\vskip -1em
\If{not \texttt{HAS(car.body, window)}}{\texttt{ADDCHILD(car.body, CarNode(window, style=1, color=None))}.} 
\end{algorithm} 
\vspace{-2em} Other possible edits to the car are \textit{remove} by deleting a specific child node of the car body, \textit{paint} by changing the color of a child node, and \textit{editpart} by changing the type of a child node. We limit the scope of the functions considered in this experiment to target only one child node of the car body at a time. The condition of all function applications are limited to targeting the same part of the car in which the function transformation takes place. The resulting functional space consisted of 1,081 distinct functions.

\subsubsection{Task procedure.} 

The experiment was split up into a training stage and a testing stage. During the training stage, participants were familiarized with 3 individual functions each represented by a different factory unit ($A$, $B$ or $C$). The functions were chosen to satisfy a set of pairwise interaction relations (Fig.\ref{fig:task}B) based on the order of application. The pair $AB$ always represents a feeding relationship, with its reverse $BA$ representing a counter-feeding relationship; the pair $CB$ always satisfies a bleeding relation, with $BC$ representing its reverse (see example trials in Fig.\ref{fig:examples}).  Eight possible function triplets were pre-generated, and each participant was randomly assigned one of the sets.

Participants observed 9 unique cars moving through each factory unit on a conveyor belt (Fig.\ref{fig:task}C). If a given car was a valid input to the function, it emerged from the unit with one part transformed; otherwise, the car remained unchanged. After the presentation of exemplars for each factory unit, a short test followed to determine whether participants inferred the expected underlying single function. Specifically, we asked them to assemble the correct output car for each of 3 prompted input cars using the game interface (Fig.\ref{fig:task}D). All example input-output pairs remained on screen throughout this period, and participants were informed of the correctness of their generations. For each input car, participants were given up to 3 opportunities to configure the correct output. 

Finally, participants were asked to generate the correct output for each prompted input car based on two different factory units shown in consecutive orders (Fig.\ref{fig:task}E). No feedback was provided during this phase. For each of the interaction pairs $AB$, $BA$, $CB$ and $BC$, participants completed a block of 8 output generations sequentially. The order of each block was randomized for each participant, and the prompted input cars were also presented in a randomized order. Throughout the testing stage, participants had access to all previously seen examples of each relevant function.

\subsubsection{Participants.} 

We recruited participants via Amazon Mechanical Turk for the online experiment. To recruit high quality participants, we first conducted a pre-experiment survey consisting of 10 simple attention-checking questions about the car assembly game setup. Of the 196 workers who responded to the survey, we invited those who scored over 90\% ($n=117$) to participate in the main experiment. Participants completed the main task within an hour and 30 minutes, and were compensated \$14.00, plus up to \$2.00 of performance-based bonus (\$2.00 $\times$ overall accuracy) upon the completion of the experiment.

\subsection{Behavioral results} 

\begin{figure*}[ht]
  \includegraphics[width=\textwidth]{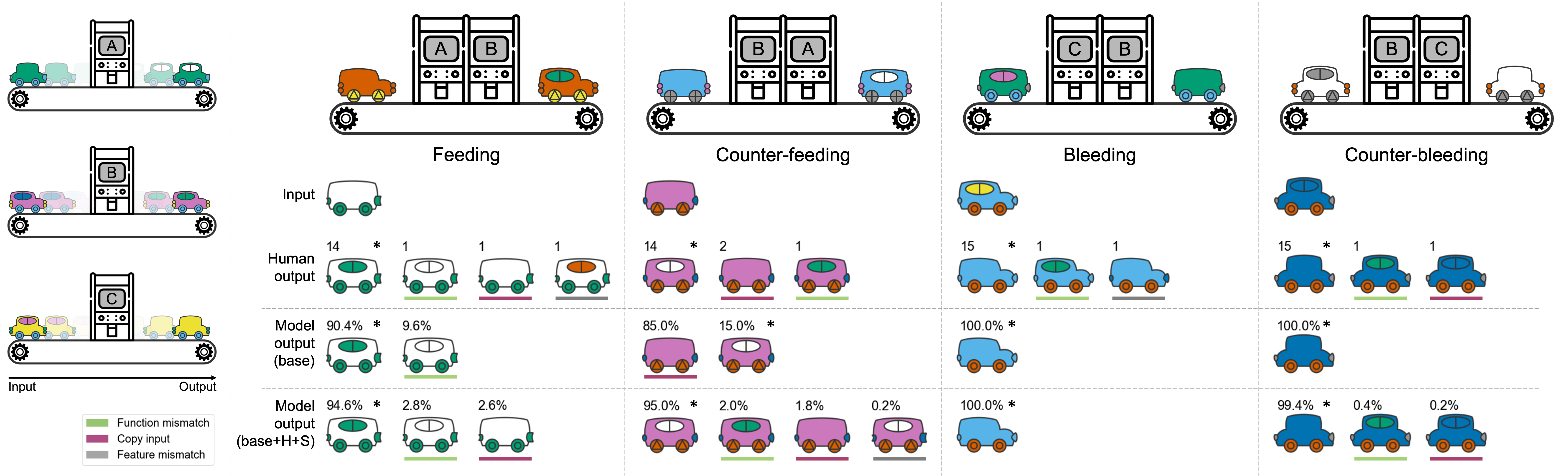}
  \caption{Examples of human and model output for each interaction type. Individual functions $A$, $B$ and $C$ are shown on the left. Prompted input cars (top rows) are shown alongside different output generations by humans, the base MLC model, and the fine-tuned MLC. The number in the top-left corner reflects the count (or percentage for model samples) of generation for each car. Correct outputs are marked by $*$; erroneous generations are underlined with colors corresponding to the error types (green: \textit{function mismatch}; purple: \textit{input copying}; grey: \textit{feature mismatch}).}
\label{fig:examples}
\vskip -1em
\end{figure*}

The first analysis examined how people performed on few-shot learning of individual, isolated functions. Specifically, during the testing stage, participants generated the expected responses for most of the prompted inputs to individual (non-compositional) functions ($M = 95.4\%$, $SEM = 0.010$). The three input cars for each single function were analyzed based on three groups: \textit{familiar}, \textit{novel}, and \textit{identity}. Cars in the \textit{familiar} group consisted of example inputs seen during the training stage, which checked for attention lapses and familiarized participants with the generation interface. The average accuracy was $95.2\%$ ($SEM = 0.012$) for this group. The \textit{novel} group contained novel cars that are valid inputs to the considered function, and the average accuracy was $95.7\%$ ($SEM = 0.012$) for this group. Finally, \textit{identity} items were invalid inputs accepted by the functions; therefore, the expected output generation for each input in this group should be an identical copy. Participants were able to identify conditions in which the input is invalid with an average accuracy of $95.4\%$ ($SEM = 0.013$) for this group. Together, high accuracy on all three groups suggests sufficient learning of the single functions on which participants were trained.

The second analysis examined how people reasoned about function composition under interaction. During test, there were four consecutive blocks, each representing one of the feeding (F), counter-feeding (CF), bleeding (BL), and counter-bleeding (CBL) interaction conditions (Fig.~\ref{fig:human_res}A). Participants achieved an average accuracy of $86.8\%$ ($SEM = 0.019$) across all trials and conditions. Additionally, we found that participants performed comparably well across interaction conditions, with an average accuracy of $85.8\%$ ($SEM = 0.028$) for the feeding trials, $86.3\%$ ($SEM = 0.027$) for counter-feeding, $86.3\%$ ($SEM = 0.024$) for bleeding and $88.8\%$ ($SEM = 0.020$) for counter-bleeding. Using test conditions as fixed effects while accounting for the random effect of participant ID, a mixed effects logistic model on generation correctness was performed. The analysis did not reveal a significant effect for each of the interaction conditions ($\beta_{BL} = -1.609, SEM = 1.111, p = 0.148$; $\beta_{CBL} = -1.904, SEM = 1.063, p = 0.073$; $\beta_{CF} = 1.109, SEM = 1.429, p = 0.438$).


We also evaluated whether biases suggested in past linguistic studies are observed in the current study in visual function learning. \citeA{Kiparsky1968} proposed the \textit{maximum utilization} bias, whereby feeding and counter-bleeding are easier to learn because both functions apply. In comparison, bleeding and counter-feeding are predicted to be harder because one function cannot apply due to a lack of context. A paired samples t-test was performed on the average accuracy for a participant for feeding/counter-bleeding trials versus for bleeding/counter-feeding trials. The results did not suggest an active maximum utilization bias ($t(116) = 0.488$, $p = 0.627$). \citeA{Kiparsky1971} also discussed a \textit{transparency} bias that favors instead the more transparent interactions of feeding and bleeding, over the more opaque interactions concerning functions that do not apply when it seems like it should (counter-feeding) or apply when it seems like it should not (counter-bleeding). Again, we did not find a significant difference in generation performance between feeding/bleeding trials and counter-feeding/counter-bleeding trials ($t(116) = -0.951$, $p = 0.343$). Overall, our results did not find any participant preferences for a particular set of function interactions in accordance with either linguistic theories.

\subsubsection{Error analysis}

For each composition of two functions, we computed the key transformation expected from input to output. For instance, composing both functions $A$ and $B$ results in the addition of a green oval-shaped window (Fig. \ref{fig:examples}). The majority of human-generated output cars were correct, and reflected the key transformation expected in all interaction conditions. The majority of incorrect generations fall into three main error types observed in the behavioral data (Fig.\ref{fig:human_res}B). The most frequent type of error, which we term \textit{function mismatch}, describes the scenario in which a participant fails to reflect the key transformation in their generation, by either applying only one of the functions to the input or by reversing the order of function application. For instance, participants sometimes only applied the first function to the input, and ignored the second function in the feeding condition, leaving the newly attached window unpainted (left-most column, green underlines in Fig.~\ref{fig:examples}). There were also cases where participants only applied the second function, as seen in the bleeding examples in Fig.~\ref{fig:examples} of cars that had a green window rather than no window. In the counter-feeding example of Fig.~\ref{fig:examples}, some participants erroneously applied function $A$ before $B$, resulting in a green window in their generations. Another type of error occurs when the participant simply makes an identical copy of the input (see examples in Fig. \ref{fig:examples}, purple underlines). In the current task setup, input-copying is sometimes indistinguishable from \textit{function mismatch} errors. For example, in the counter-feeding examples in Fig. \ref{fig:examples} that are purple-underlined, it is unclear whether participants intended to explicitly copy the input as a strategy, or whether they tried applying only the first function, ending up with an identical copy due to input invalidity. Finally, in the \textit{feature mismatch} scenario, participants correctly reflected the key transformation but made a mistake in a part of the car irrelevant to the functions in Fig. \ref{fig:examples}, grey underlines).

\section{Composing functions with meta-learning}

To develop a computational account of human behavior in the function composition task, we trained a neural network for learning compositional functions following the meta-learning for compositionality approach (MLC) set forth by \citeA{Lake_Baroni_2023}. Similar to the motivation in \shortciteA{mccoy2023}, instead of modeling the learning process, we aim to model human inductive biases by forming a generative process over systems of functions (equivalent to a Bayesian prior) and using samples from this process as meta-learning episodes for the neural network. We then further enrich the network with additional behavioral nuance by fine-tuning the model on behaviorally-guided data distributions \shortcite{Lake_Baroni_2023,zhou2024}.


\begin{figure}[h]
\includegraphics[width=1.0\linewidth]{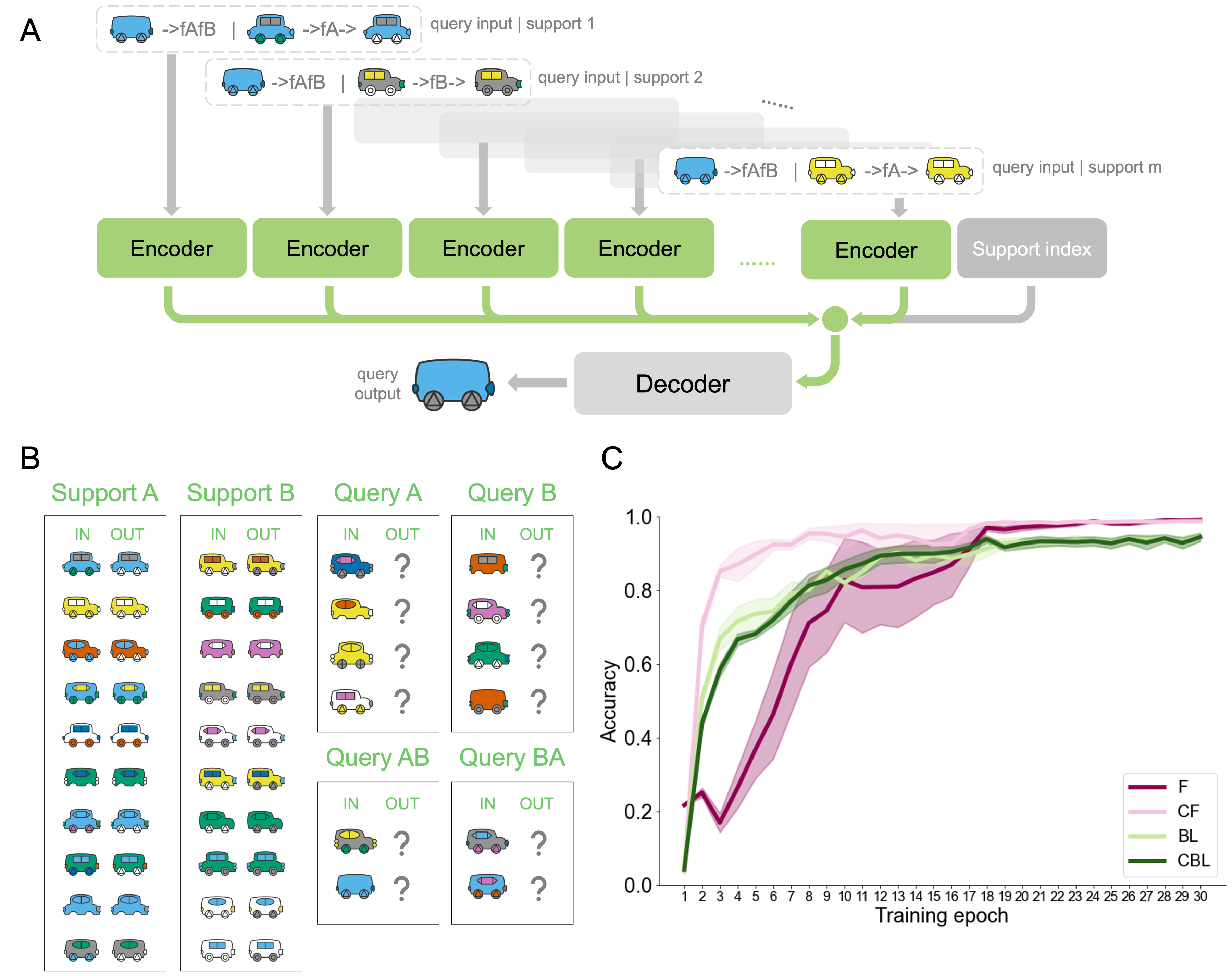}
  \caption{Model schematics and training details.}
\label{fig:model}
\vskip -1.5 em
\end{figure}

\subsubsection{Model description.}

A schematic of the MLC model structure is in Fig.~\ref{fig:model}A. The MLC model uses a standard sequence-to-sequence Transformer optimized to generate the correct output in response to a novel query based a set of support examples \cite{Lake_Baroni_2023}. To process the study examples and the queries, an encoder network takes as input a sequence of strings each representing the input car, the function handle and the corresponding output car for each support example (e.g., $\text{car}_{in} \rightarrow  fA \rightarrow  \text{car}_{out}$), and only the input car and function handle(s) for each query (e.g., $\text{car}_{in}  \rightarrow  fAfB  \rightarrow $). The tree structure of each car object is flattened into a string representation to form each $\text{car}_{in}$ and $\text{car}_{out}$ (see an example in Fig.\ref{fig:task}A), before concatenation with function handles indicating the participating functions and the ``$\rightarrow$" separator in between. Each query is duplicated and paired with each of the $m$ study examples per episode forming $m$ input sequences, which are processed by the shared encoder into $m$ separate latent embeddings. The set of latent embeddings are then each combined with an index embedding indicating its original place in the support set, before being processed together by a decoder network that generates the output sequence $\text{car}_{out}$. The encoder and the decoder are both 2-layer Transformers with 8 attention heads in each layer; additional training hyper-parameters are set to default values used in \citeA{Lake_Baroni_2023}.

\subsubsection{Base training.} 

To simulates the level of adult skills on composing functions when coming into the experiment, we train the MLC model via meta-learning \cite{Hospedales2022}, encouraging the model to develop task-general compositional skills through a series of episodes that each constitutes a different function composition task. The objective of each task is to generate the correct output of a novel composition of two functions, given the input and a limited set of support input-output examples for each function. Each episode consisted of (1) support input-output pairs of two functions $A$ and $B$; (2) A set of query inputs for the single functions $A$ and $B$, as well as the compositions $AB$ and $BA$ (see Fig.\ref{fig:model}B). This setup is comparable to the behavioral experiment, where participants were only trained on two functions and tested on their compositions. 

To form the episodes used in base training, we first divided all possible 1,081 car-edit functions into a training set and a held-out validation set (with no individual functions appearing in both). We also withheld all functions used in the behavioral experiment from the training. To generate each training episode, we randomly sampled two functions (arbitrarily labeled $A$ and $B$) from the training distribution that either satisfy a feeding/counter-feeding relationship or a bleeding/counter-bleeding relationship. For each function, we sampled 6 valid support input-output examples, and 4 invalid examples (input and output are identical). To train the model on generalizing to both new input of learned single functions, as well as novel function compositions and interactions, we queried the model with (1) novel inputs for each single function (e.g., $\text{car}_{in}  \rightarrow  fA  \rightarrow $), of which 2 are valid and 2 are invalid inputs; (2) inputs to function compositions (e.g., $\text{car}_{in}  \rightarrow  fAfB  \rightarrow $), two for $fAfB$, and two for $fBfA$. All query items were randomly sampled from all possible inputs to the relevant function(s) given validity constraint. We generated 50,000 training episodes in total, half of which contained feeding/counter-feeding function pairs and the other contained bleeding/counter-bleeding functions. With a batch size of 5 episodes, the model parameters were optimized via minimizing the average cross-entropy loss per output token over 50 epochs. Over the course of training, the model must learn to represent single functions based on the support examples, and to compose two functions according to the order of function application, all without explicit compositional supervision. 

\subsubsection{Behaviorally-informed fine-tuning.} 

To capture additional behavioral nuance, we fine-tuned the MLC model on two data distributions reflective of the human experience. The first data distribution $H$ was formed using raw human data. We used human responses on 4 out of the 8 total function triplets assigned to participants in the behavioral experiment for additional training, and reserved the rest for evaluation. Each episode in $H$ was structured identically to the base training episodes, except that support input-output pairs and queried items were replaced with the ones used in the behavioral experiment. Since participants were trained on 3 single functions, each participant's data was used to form two separate episodes, one for feeding/counter-feeding trials, and the other for bleeding/counter-bleeding trials. As a result, the $H$ distribution contained 110 episodes, each with 18 support examples (9 for each function), 6 single function queries (3 for each function) and 16 composition queries (8 for each function composition). Additionally, we created a synthetic data distribution $S$ containing 2,000 episodes that is human behaviorally-guided. Specifically, we modified the generative process for creating the base training distributions such that noise was injected into the output sequences to the queries by flipping the function order with a small probability $p_{flip}$. This simple manipulation creates similar error patterns in output generations like the ones underlined in green and purple shown in Fig.~\ref{fig:examples}. Together, the model parameters were updated to predict the raw human generations in $H$ and noisy outputs in $S$ during the fine-tuning stage, instead of the ground-truth sequence.

\subsection{Modeling results}
 Using the reserved validation function set, we created a set of 3,000 base validation episodes. At the conclusion of base training, the MLC model generated the correct output sequence for each query at an average of $97.9\%$ accuracy across query types and different random initializations ($SEM = 0.037$). When queried on single functions, the model was able to make near-perfect generalizations ($M = 98.1\%$, $SEM = 0.004$). When divided into interaction types, validation accuracy on function compositions remained consistently high ($M_{F} = 97.6\%$, $SEM_{F} = 0.007$; $M_{CF} = 98.0\%$, $SEM_{CF} = 0.004$; $M_{BL} = 97.3\%$, $SEM_{BL} = 0.004$; $M_{CBL} = 97.1\%$, $SEM_{CBL} = 0.005$).

\begin{figure}[h]
\includegraphics[width=1.0\linewidth]{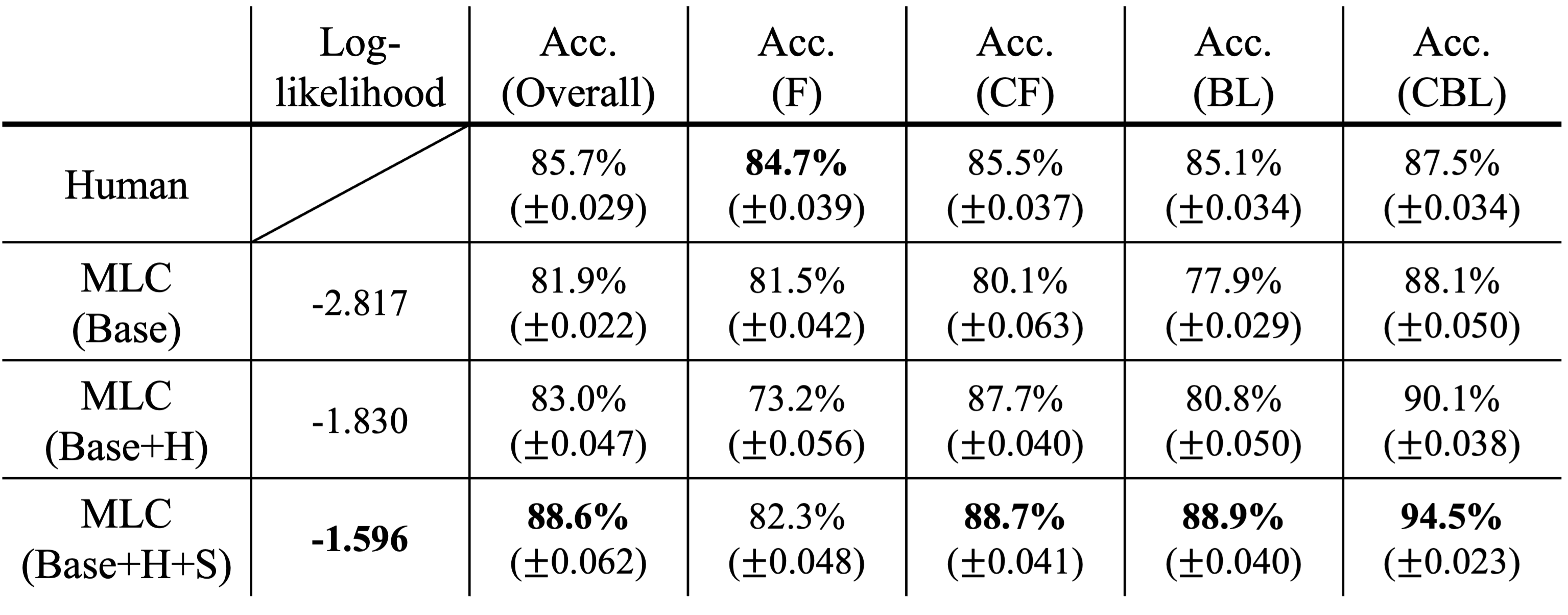}
  \caption{Modeling results. (Column 1) Model goodness of fit for predicting human generated examples. For each model version, the overall average log-likelihood per human generated output car is in the first column. (Column 2-6) Human vs. model accuracy by interaction type on held-out experiment trials. Bolding indicates best performance.}
\label{fig:table}
\vskip -1.5em
\end{figure}

To directly compare the generalization behavior of the model with humans, we evaluated generations from MLC models on the set of experimental trials using the 4 held-out function triplets. Model accuracy results are summarized in Fig.~\ref{fig:table}. We observed slightly lower overall accuracy from the MLC model with only base training. However, our primary modeling goal here is not to train a model that surpasses humans on these trials but to provide a comprehensive account of human compositional generalization behavior. Reviewing the examples of model samples shown in Fig.~\ref{fig:examples}, we observe that the base MLC model can produce the correct output most of the time, but the samples do not reflect the variety of human-like error patterns. With additional behavioral guidance by fine-tuning on the $H$ and $S$ distributions, we observed an improved account of human behavior in terms of both accuracy and average log-likelihood for held-out human data (Fig.~\ref{fig:table}). When fine-tuned with only human-generated data ($H$), with a minimal number of training episodes compared to the base training set, the MLC model already demonstrated an improved account of held-out data in terms of log-likelihood. When MLC model was provided with additional guidance from both $H$ and $S$, it performed the best on held-out data and exhibited an overall improvement in log-likelihood over the base MLC model. Qualitatively, examples generated by the fine-tuned MLC model show a more diverse set of error patterns mirroring human data Fig.~\ref{fig:examples}, demonstrating the effectiveness of the additional behavioral enrichment.


\section{Discussion}

To study zero-shot visual function composition in both humans and machines, we designed a learning paradigm to examine generation behavior under different interaction conditions. Specifically, through training participants on small sets of input-output examples, we observed highly accurate generalizations to novel function inputs. During the test phase, we presented participants with queries involving the four logical patterns that result from combining pairs of interacting functions. Contrary to previous work in a different domain which finds different accuracy across interaction types, our results indicate consistently high levels of generation accuracy across all conditions. This suggests humans can adeptly handle contextual shifts when composing functions. While we do not observe overall mean differences between different interaction types, we observed that humans make non-random, structural mistakes when processing two functions consecutively: partial application of functions or a reversal of function order constitute the main families of errors.

In a side-by-side evaluation of human and machine performance, we found that a standard sequence-to-sequence Transformer can be trained to compose novel function pairs via meta-learning on streams of function composition tasks. In particular, our MLC model was capable of learning both single functions from limited examples and generalizing to unseen function composites by producing output sequences to promoted inputs at a near-human accuracy. Additionally, we used error patterns uncovered in behavioral data analysis to form the basis of comparison with model error patterns, and further revealed how distant models are from humans in understanding function interactions. To decrease the gap in human-model generalization behavior, we conducted an additional experiment by fine-tuning the model on behaviorally-informed data distributions, and the results indicated that the model can be further improved to provide better accounts of human behavior within the existing framework.

Although we found relatively uniform levels of performance across conditions in both behavioral and computational analyses, the various learning trajectories of the 4 function interaction types during base model training might shed light on how different processes are mastered over the course of development (Fig.\ref{fig:model}C). While validation performance ultimately converged to high levels at the end of training, feeding was clearly the hardest for the model to master, followed by bleeding and counter-bleeding, with counter-feeding learned most early and easily. This result was expected as feeding always requires the model to apply two function transformations when the input is valid, while in other cases only one function application takes place at times. However, it is interesting to note that counter-feeding, or the application of only the second function in a novel composite, is also demonstrated to be an infant behavior before the onset of function composition skills \cite{Piantadosi2018}. Extending the current study to evaluate populations at various developmental stages might help elucidate if certain function interaction types are indeed harder to grasp. 

As a future step, sequences currently used by our model can be readily translated into other representational formats such as raw images, which are what human participants observed, or tree structures, which do not enforce arbitrary ordering of car parts and features.  More direct comparisons between human and model behavior can offer additional insight on how function interactions are learned and processed, and may further inform how to build computational systems with more compositional, human-like forms of learning.

\section{Acknowledgements}
This work was supported by NSF Award 1922658 NRT-HDR: FUTURE Foundations, Translation, and Responsibility for Data Science. Yanli Zhou was supported by the Meta AI Mentorship Program. We thank Solim LeGris and Cindy Luo for helpful discussions of this manuscript.

\bibliographystyle{apacite}

\setlength{\bibleftmargin}{.125in}
\setlength{\bibindent}{-\bibleftmargin}

\bibliography{library_clean}

\end{document}